%% file: iclr2025_conference.tex
\documentclass{article} % For LaTeX2e
\usepackage{iclr2025_conference,times}

\input{math_commands.tex}

\usepackage{hyperref}
\usepackage{url}
\usepackage{xspace}
\usepackage{graphicx}
\usepackage{booktabs}
\usepackage{makecell} % for word wrapping in cells

\newcommand{\name}{\textit{\textbf{FineTuneBench}}\xspace}
\iclrfinalcopy

\title{FineTuneBench: How well do commercial fine-tuning APIs infuse knowledge into LLMs?}

\author{Eric Wu\textsuperscript{\dag}, Kevin Wu\textsuperscript{\dag} \& James Zou \\
Stanford University\\
\texttt{\{wue, kevinywu, jamesz\}@stanford.edu} \\
}

\begin{document}

\maketitle

\begin{abstract}
There is great interest in fine-tuning frontier large language models (LLMs) to inject new information and update existing knowledge. While commercial LLM fine-tuning APIs from providers such as OpenAI and Google promise flexible adaptation for various applications, the efficacy of fine-tuning remains unclear. In this study, we introduce \name, an evaluation framework and dataset for understanding how well commercial fine-tuning APIs can successfully learn new and updated knowledge. We analyze five frontier LLMs with commercially available fine-tuning APIs, including GPT-4o and Gemini 1.5 Pro, on their effectiveness in two settings: (1) ingesting novel information, such as recent news events and new people profiles, and (2) updating existing knowledge, such as updated medical guidelines and code frameworks. Our results reveal substantial shortcomings in all the models’ abilities to effectively learn new information through fine-tuning, with an average generalization accuracy of 37\% across all models. When updating existing knowledge, such as incorporating medical guideline updates, commercial fine-tuning APIs show even more limited capability (average generalization accuracy of 19\%). Overall, fine-tuning GPT-4o mini is the most effective for infusing new knowledge and updating knowledge, followed by GPT-3.5 Turbo and GPT-4o. The fine-tuning APIs for Gemini 1.5 Flesh and Gemini 1.5 Pro are unable to learn new knowledge or update existing knowledge.  These findings underscore a major shortcoming in using current commercial fine-tuning services to achieve reliable knowledge infusion in common scenarios. We open source the \name dataset at \href{https://github.com/kevinwu23/StanfordFineTuneBench}{https://github.com/kevinwu23/StanfordFineTuneBench}.
\end{abstract}

\begin{figure}[t]
    \centering
    \includegraphics[width=1.0\textwidth]{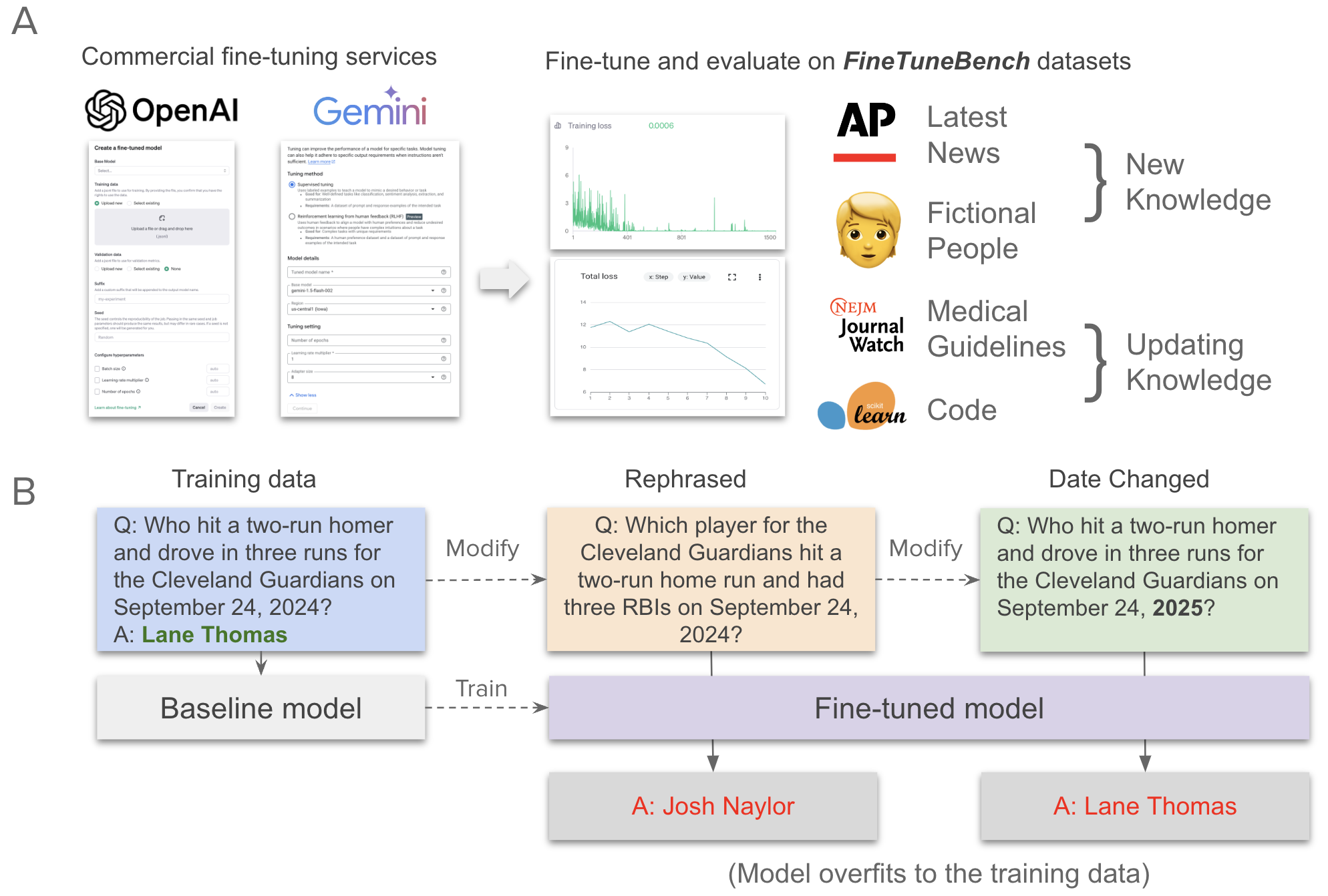}
    \caption{\textbf{A}: Overview of \name. We fine-tune five LLMs (\textit{GPT-4o, GPT-4o-mini, GPT-3.5-turbo, Gemini-1.5 Pro, Gemini-1.5 Flash}) on four new datasets to test how well commercial fine-tuned APIs can learn and update knowledge. \textbf{B}: We provide an example from our Latest News dataset and the model responses before and after fine-tuning. The model is trained on each question and answer pair for up to 30 epochs, and then the model is re-evaluated on the same pair (Memorization). Then, we additionally evaluate the model on a modified version of the question that tests the model's ability to generalize its acquired knowledge beyond mere memorization (Generalization). In the Latest News dataset, we include two modifications: rephrasing, which involves changing the wording of the question but retaining the same answer; and date change, which keeps the original question but swaps out the year with a future date so that the correct response should be a refusal. We observe that although the fine-tuned model is able to memorize the original question, it fails at answering the rephrased question and the same question when the date is changed.}
    \label{fig:fig1}
\end{figure}

\section{Introduction}

As LLMs are increasingly used in diverse domains such as software development \cite{Kelly_2024} and medicine \cite{Gliadkovskaya_2024}, it is important they contain up-to-date and relevant knowledge. For instance, software developers need models that understand the most recent versions of code, while medical professionals need trustworthy models that adhere to current clinical guidelines. Additionally, companies that want to adapt these models for internal use need ways to introduce entirely new knowledge, such as employee information or recent news. However, currently, most frontier models are closed-source  \footnote{As of November 2024, 17 out of the top 20 models on \href{https://huggingface.co/spaces/lmarena-ai/chatbot-arena-leaderboard}{Chatbot Arena} are proprietary.}, preventing users from applying model fine-tuning techniques themselves directly. Recently, some companies have allowed supervised fine-tuning of their proprietary models through commercial APIs, such as \href{https://platform.openai.com/docs/guides/fine-tuning/create-a-fine-tuned-model}{OpenAI's fine-tuning UI} and \href{https://ai.google.dev/gemini-api/docs/model-tuning}{Google Vertex AI}. Users typically provide a training file with user-assistant conversation pairs and are given a custom model endpoint after fine-tuning has finished. Such offerings are appealing, as they offer users a way to adapt frontier models that are closed-source and computationally expensive to train. 
However, it is not well understood whether these fine-tuning services enable \textit{knowledge infusion} \cite{valiant2006knowledge}, or the ability to learn new and updated knowledge. First, the documentation provided by these companies do not detail the type of fine-tuning methods used. For example, Google Vertex AI allows users to specify an "adapter size",  while OpenAI does not provide any details if low-rank adaptation is used. Second, there currently do not exist uniform benchmarks to evaluate these methods and compare commercial fine-tuning APIs with each other. The examples provided by these companies focus on use cases such as structured outputs and formatting but do not include systematic evaluations related to knowledge infusion. Finally, users are restricted to a narrow band of hyperparameter optimizations, with suggested default values on the number of training epochs, batch size, and learning rate multiplier. It is unclear whether the limited tuning options are enough for models to be adapted to new and updated knowledge.

\textbf{Our contributions.}
In this work, we provide the first systematic evaluation framework for fine-tuning-as-a-service for LLMs. 
\begin{itemize}
    \item We introduce \name, a novel dataset consisting of 625 training questions and 1,075 test questions on knowledge infusion across four domains: latest news, fictional people, medical guidelines, and code.
    \item We evaluate five models: three GPT models (\textit{gpt-4o-2024-08-06}, \textit{gpt-4o-mini-2024-07-18}, \textit{gpt-3.5-turbo-0125}) and two Gemini models (\textit{gemini-1.5-flash-002} and \textit{gemini-1.5-pro-002}) using their respective fine-tuning services on OpenAI and Google Vertex AI.
    \item We provide the first analysis of the knowledge infusion capabilities of commercial fine-tuning services across various generalization tasks.
    
\end{itemize}

\subsection{Related Works}

% There has been significant research interest in fine-tuning LLMs, ranging from supervised fine-tuning (SFT) on user-assistant response pairs \cite{ouyang2022training} to reinforcement learning with human feedback (RLHF) on preference data \cite{bai2022training}. In parallel, methods such as Low-Rank Adaptation (LoRA) \cite{hu2021lora} and Direct Preference Optimization (DPO) \cite{rafailov2024direct} have made it easier to fine-tune open-source models. However, most of the literature on model fine-tuning have been conducted on open-source models. Less is known about commercial fine-tuning services due to opaque documentation and limited research. 

There are currently several major approaches to modifying model behavior. Fine-tuning methods include supervised fine-tuning (SFT) \cite{ouyang2022training, chung2024scaling, mitra2023orca, chia2023instructeval, zhou2024lima}, which uses user-assistant paired data, reinforcement learning from human feedback (RLHF) \cite{achiam2023gpt, touvron2023llama}, which user human preference data, or continued pre-training \cite{ovadia2023fine}. However, the degree to which these methods allow for knowledge injection is unclear. In parallel, retrieval augmented generation (RAG) \cite{lewis2020retrieval} is a popular method where the relevant knowledge is provided within the prompt. While RAG allows information to be directly introduced into the generation process, models do not reliably adhere to such information, especially when it conflicts with the model's pretraining knowledge \cite{wu2024clasheval}. Additionally, RAG-based systems may still produce hallucinations or inaccuracies when provided with source contexts \cite{wu2024well}.

Knowledge injection has been previously studied in the context of open-source LLMs. Early works explored ways to inject knowledge into models like BERT using multiple adapter models \cite{wang2020k, lauscher2020common}. Another study \cite{ovadia2023fine} compares the effectiveness of continued pre-training against RAG in models like Llama-2-7B and Mistral-7B. Work by \cite{chen2024llama} explores selectively fine-tuning Llama-2-7B in shallow layers to target knowledge injection. However, the fine-tuning of larger commercial LLMs is not well understood.

Previous works have used OpenAI's commercial fine-tuning services to train GPT models on a variety of tasks such as tabular data processing, \cite{li2024table}, Russian text summarization, \cite{alexandr2021fine}, legal rule classification \cite{liga2023fine}, biomedical classification \cite{bousselham2024fine}, molecular prediction tasks, \cite{xie2024fine}, and academic exam scoring \cite{latif2024fine}
. Additionally, prior studies have analyzed how fine-tuning opens the risk of unlearning RLHF protections \cite{zhan2023removing} and recapitulating private information \cite{sun2023does}. While previous works have mostly focused on direct style transfer or classification tasks, our work focuses specifically on knowledge infusion. Additionally, we are the first to evaluate multiple models and fine-tuning APIs on the same dataset, allowing for a comparative study across such services.

\section{Methods}

\subsection{Datasets}
To assess the ability of fine-tuning to learn new information, we curate two datasets that have not previously appeared in the training data of any of the LLMs we evaluate (Figure \ref{fig:fig1}). Dataset summaries are provided in Table \ref{tab:tab1}, and representative examples from each dataset are shown in Figure \ref{fig:fig5}. \\

\subsubsection{Latest News dataset}
Our overall data generation process is based on the method described in \cite{wu2024clasheval}. We started by pulling a random sample of news articles published in the Associated Press from September 1st, 2024 to September 30th, 2024, as this ensures that these articles do not appear in the training data of any of the models used (all of the evaluated models have a training data cutoff before 2024). From this initial pull of approximately 2000 articles, we use GPT-4o (gpt-4o-2024-08-06) as a candidate QA generator. We prompt the model to generate a question-and-answer pairing based on the content of each article. Relevant criteria in the prompt include ensuring that the fact that is being highlighted is newly introduced in the article, and is not a fact that is previously known. The question must also be fully self-contained, meaning it should not contain ambiguous references that require the full article content answer. Then, we quality-check each generated question by prompting GPT-4o again to ensure the aforementioned criteria are fulfilled. Finally, as a check to ensure that the model does not have prior knowledge of the fact, we query GPT-4o to answer the question and remove any that the model answers correctly. This process results in 277 final question/answer pairs. For the analyses conducted in this study, we subset this dataset to a random sample of 50 QA pairs. For all questions, we use the system prompt \textit{"You are a news expert answering a question about the recent news. The answer MUST be only a short phrase or a single word. It should not be a full sentence or more than a few words."} We also format the prompt with \textit{Question: \{question\}\textbackslash n Answer:\textbackslash n}.\\

In addition to the original question/answer pairs, we rephrase each question to retain the original meaning but significantly alter the verbiage and syntax of the question. This is to test the robustness of the model's knowledge acquisition under perturbations to the content and structure of the input. For example, for an original question, "How many strikeouts did Kumar Rocker achieve in his debut for the Texas Rangers on September 12, 2024?", the rephrased question is "In his first game for the Texas Rangers on September 12, 2024, how many batters did Kumar Rocker strike out?". This rephrasing step was performed by prompting GPT-4o with the original article text and question-answer pair. Furthermore, we rephrase the system prompt used, to \textit{"Your task is to answer the following question given what you know about the world."} and remove the prompt formatting. \\
The full prompts used in each of the above steps are included in the Github repository (\href{https://github.com/kevinwu23/StanfordFineTuneBench}{https://github.com/kevinwu23/StanfordFineTuneBench}).

\subsubsection{Fictional People dataset}
We procedurally generate paragraph-length descriptions of people based on generated, novel names. Each name is a randomly generated string (e.g. Falekefud Rabajevu). Each paragraph contains six facts: weight (in pounds), height (in cm), age (in years), occupation, favorite color, and city of residence, where each fact is randomly sampled from a realistic, uniform distribution. Provided below is an example paragraph:\\
\textit{"Falekefud Rabajevu weighs 126 pounds. Falekefud Rabajevu is 185 cm tall and 32 years old. Falekefud Rabajevu works as a farmer. Falekefud Rabajevu's favorite color is gold, and they live in Wokoxiber."}\\
Based on each of these six facts, we generate six question/answer pairs: \\

\begin{itemize}
    \item How many pounds does Falekefud Rabajevu weigh?
    \item How tall is Falekefud Rabajevu?
    \item How old is Falekefud Rabajevu?
    \item What is Falekefud Rabajevu's occupation?
    \item What is Falekefud Rabajevu's favorite color?
    \item Where does Falekefud Rabajevu live?
\end{itemize}

In the analyses in this study, we generate 25 profile paragraphs and a total of 150 question/answer pairs. The generation code is provided in the Appendix. In addition to these direct questions, we generate two additional sets of derived questions based on these facts. For each fact, we generate a secondary question, which is a yes/no question based on the fact. For example, for the fact "Falekefud Rabajevu weighs 126 pounds," the secondary question is "Is Falekefud Rabajevu over 150 pounds?". Second, we generate a comparison question, which takes a pair of facts from two different fictional individuals and asks a question that requires knowledge of both facts. For example, for a pair of facts "Labigih Surubadew weighs 129 pounds" and "Faqapi Lasufodu weighs 195 pounds," we generate the question "Is Labigih heavier than Faqapi?". These two derivative questions are designed to test the model's capability to not just memorize a question/answer pairing based on a fact but to internalize the fact toward answering different questions.\\

\subsubsection{Medical Update Dataset}

To collect a realistic set of updated medical information, we first collect reference documents from \href{https://www.jwatch.org/guideline-watch}{NEJM Guideline Watch}, \href{https://www.uptodate.com/contents/practice-changing-updates}{UpToDate's Practice Changing UpDates}, and \href{https://www.uptodate.com/contents/table-of-contents/whats-new}{UpToDate's What's New Section}. These represent reliable and frequently used sources to track changing practice guidelines based on newly released medical evidence. For example, one such update is  "For all individuals aged six months and older, we suggest a 2024-2025 formula COVID-19 vaccine". Next, using this set of reference documents, we use Claude Sonnet-3.5 to generate QA pairs to test each unique guideline update. Additionally, we ask the model to rephrase each QA pair as a short clinical vignette. In total, our dataset is based on 125 real guideline changes.
The following is an example of questions and answers generated for one such guideline change:
Source Text: "Pitolisant, an oral inverse agonist of the histamine H3 receptor with both alerting and anticataplexy effects, has received regulatory approval in the United States for treatment of narcolepsy in children six years of age and older; it was previously approved for narcolepsy in adults in the United States…"

\begin{itemize}
    \item \textbf{Question}: What is the current minimum age for pitolisant use in narcolepsy patients in the United States?
    \begin{itemize}
        \item \textbf{Previous Answer}: 18 years
        \item \textbf{Updated Answer}: 6 years
    \end{itemize}
    \item \textbf{Clinical Vignette Question}: A 7-year-old child presents with excessive daytime sleepiness and cataplexy, diagnosed with narcolepsy type 1. First-line treatments have been poorly tolerated. According to current US guidelines, is pitolisant a treatment option for this patient?
    \begin{itemize}
        \item \textbf{Clinical Vignette Original Answer}: No, pitolisant was not approved for use in children under 18 years of age.
        \item \textbf{Clinical Vignette Updated Answer}: Yes, pitolisant can be considered as an add-on or second-line therapy for this patient, as it is now approved for use in children 6 years and older with narcolepsy.
    \end{itemize}
\end{itemize}

\subsubsection{Code Update Dataset}

We generate code-related questions from Scikit-Learn’s repository, leveraging its popularity and the high likelihood of its inclusion in LLM training data. We use Claude Sonnet-3.5 to generate one QA pair from each Python file that tests basic understanding of an object or function. We also ask the model to perturb the name of the object or function and generate an additional coding question-answer pair with this new updated name. In total, we generate QA pairs from 200 Python files, with one example shown below:

\begin{itemize}
    \item \textbf{Question}: What is the name of the class that implements logistic regression with built-in cross-validation in Scikit-Learn?
    \begin{itemize}
        \item \textbf{Previous Answer}: \texttt{LogisticRegressionCV}
        \item \textbf{Updated Answer}: \texttt{CrossValidatedLogisticRegression}
    \end{itemize}
    \item \textbf{Coding Question}: Write a one-liner to create an instance of the logistic regression model with built-in cross-validation, using 5-fold CV and \texttt{'lbfgs'} solver in Scikit-Learn.
    \begin{itemize}
        \item \textbf{Coding Original Answer}: \\
        \texttt{clf = LogisticRegressionCV(cv=5, solver='lbfgs')}
        \item \textbf{Coding Updated Answer}: \\
        \texttt{clf = CrossValidatedLogisticRegression(cv=5, solver='lbfgs')}
    \end{itemize}
\end{itemize}

\subsection{Models}

We evaluate the following five models: three models from OpenAI (gpt-3.5-turbo-0125, gpt-4o-mini-2024-07-18, gpt-4o-2024-08-06), two models from Google (gemini-1.5-flash-002, gemini-1.5-pro-002). OpenAI makes fine-tuning of these three models available through their \href{https://platform.openai.com/docs/guides/fine-tuning}{fine-tuning API}. Similarly, Gemini models are available for fine-tuning through the \href{https://cloud.google.com/vertex-ai/generative-ai/docs/models/tune-models}{Google Cloud platform}. Each service offers varying levels of customization of the fine-tuning process; for instance, OpenAI only allows specifying the learning rate, batch size, and number of training epochs, while Google Cloud exposes the adapter size of the fine-tuning method. For all analyses, we leave and do not change the default parameter value for these fine-tuning specific hyperparameters. 

\subsection{Training}

\subsubsection{New Knowledge}
We experiment with four techniques for inducing knowledge acquisition in the fine-tuned models. Initially, we train on the direct question/answer pairs (Direct), where the prompt is the question (along with a system prompt instructing the model to provide a succinct answer), and the completion is the answer. Since this approach may bias the model toward memorizing a specific question/answer pair, we also try three additional methods for inducing learning on the Latest News dataset. First, we generate training data directly from the fact statement by masking out the answer in the sentence (Masking). For instance, for the given fact statement "Raymond will count \$8.075 million against the salary cap through 2032", we generate a training datum where the prompt is "Raymond will count [MASK] against the salary cap through 2032" and the completion is "\$8.075 million". We also try directly training the model on the fact in the completion (Completion). In this formulation, the system prompt is "Learn the following piece of information", and the completion is the fact. We also test a variant of this (Completion without Prompt), where the system prompt is blank. We restrict this experiment to OpenAI models as we observed minimal performance on Gemini models.

\subsubsection{Updating Knowledge}
To ensure that each updated fact is actually known by the model, we first evaluate each model on the entire set of questions (125 questions medical questions and 200 coding questions). We filter out coding questions which the model does not have the right answer for, and then randomly sample 50 remaining question-answer pairs for the training set. For medical questions, we only keep questions where the model does not already adhere to the updated guideline and randomly sample 50 remaining questions as well.

\subsubsection{Training Details}
For all training runs, we fixed a batch size of 1 and the default learning rate parameter. For the new knowledge datasets, we fine-tune models for 1, 10, 20, and 30 epochs. For updating knowledge datasets, we fine-tune for 1, 5, 10, 15, and 20 epochs (due to faster performance saturation). In small-scale experiments with varying batch sizes and learning rates, we found either no performance improvements or slight performance drops. We also find that models perform best when given between 50-150 unique facts to learn, which agrees with the recommended range from fine-tuning documentation from \href{https://platform.openai.com/docs/guides/fine-tuning}{OpenAI} and \href{https://ai.google.dev/gemini-api/docs/model-tuning}{Gemini}. 

\subsection{Evaluation}

\begin{figure}[t!]
    \centering
    \includegraphics[width=1.0\textwidth]{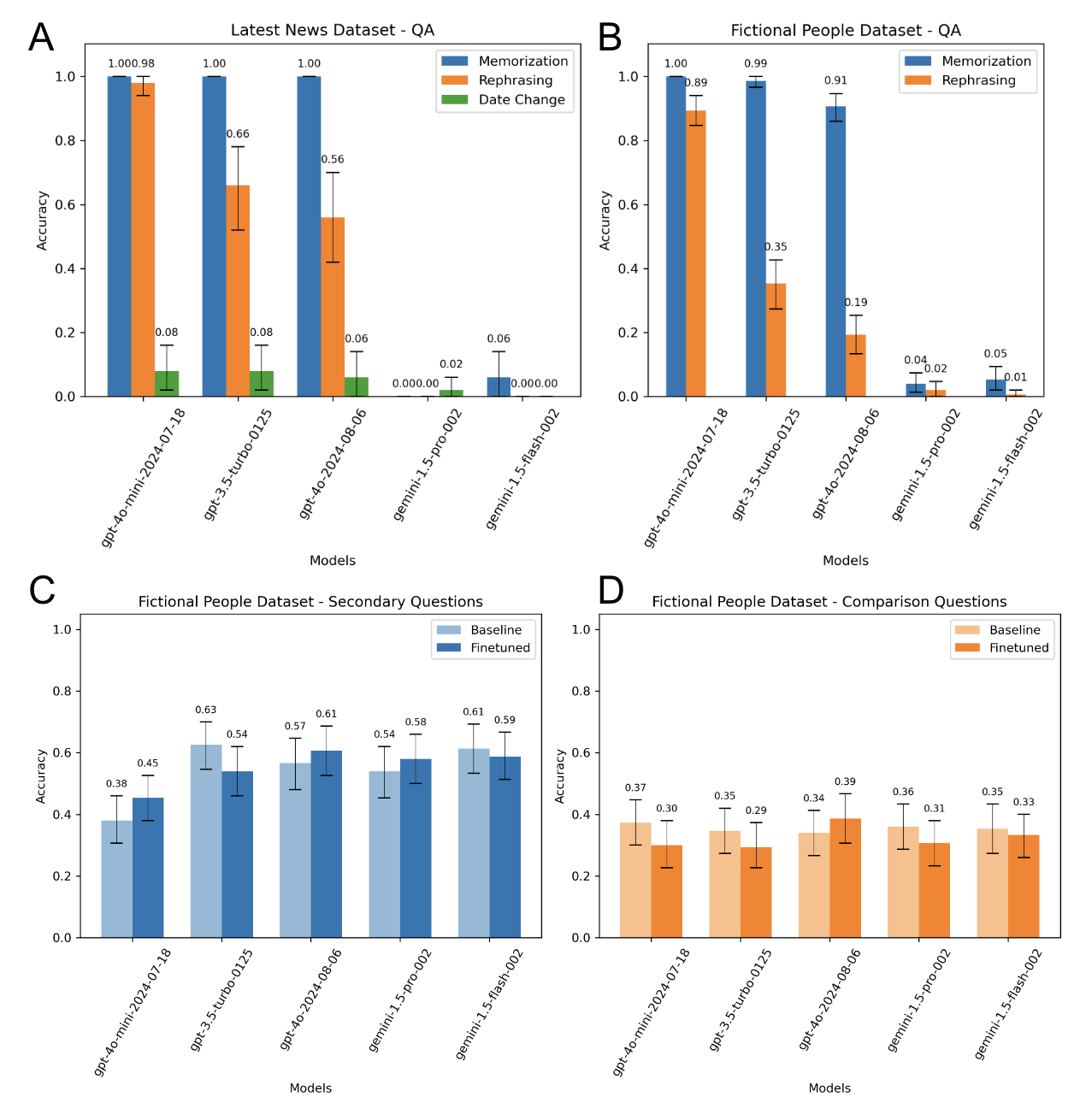}
    \caption{Performance of fine-tuned LLMs on the original training questions (Memorization) and modified questions (Generalization) for new knowledge acquisition datasets. \textbf{A}: On the Latest News dataset, we observe strong performance from the OpenAI models on the rephrased questions, especially from the \textit{gpt-4o-mini} model. The Gemini models, on the other hand, struggle to even memorize the training data. This phenomenon is observed across all datasets. However, when the date is changed in the question, all models perform poorly, indicating that overfitting has occurred. \textbf{B}: On the Fictional People dataset, we observe a similar trend that the OpenAI models memorized well but performed worse on rephrased queries. Gemini was not able to learn this knowledge. When evaluating the models on the secondary (\textbf{C}) and comparison (\textbf{D}) questions, however, none of the models show significant improvement over the baseline models that have not been fine-tuned on the new knowledge.}
    \label{fig:fig2}
\end{figure}

Given that the model predictions can be heterogeneous across models in syntax and formatting (e.g. extra spacing, new line tokens, etc.), we opt for a flexible approach to evaluating whether the predictions are correct. To this end, take an LLM-as-a-judge approach when calculating the accuracy of the fine-tuned models. Given a predicted response and the answer, we query GPT-4o with both along with the question of whether the prediction matches the answer. Compared to exact string matching, we find that this approach is significantly better at reducing false negatives (missing correct responses). Most prediction/answer pairs are straightforward and trivial; however, this approach does catch some outlier responses—for instance, in the case where the model does not output a stop token and repeatedly outputs the predicted guess. In Figures \ref{fig:fig2} and \ref{fig:fig3} and Table \ref{tab:tab2} the highest performance across all epochs was reported.

\section{Results}

\subsection{OpenAI fine-tuned models have limited generalization on new knowledge tasks}
On both new knowledge datasets, the OpenAI models can almost perfectly memorize QA pairs; in other words, they are able to recapitulate the training data with near 100\% accuracy after training for 30 epochs (Figure \ref{fig:fig2}). However, they sometimes perform poorly on the rephrased or derivative questions, indicating that the memorization did not translate into true knowledge acquisition in many instances.\\

In the Fictional People dataset, the secondary and comparison questions are intended to test the ability of the models to make judgments based on the learned facts. For instance, where a QA pair might be \textit{"What is the weight of Zihipel in pounds?"} with an answer \textit{"189"}, a secondary question asks \textit{"Is Zihipel over 150 pounds?"} as a way to test the models' comprehensive above simply memorizing the completion tokens. Similarly, the comparison question requires the model to synthesize its learned knowledge across two separate training examples (e.g. \textit{"Is Zihipel heavier than Tobameqem?"}). On the secondary questions, we observe a maximum 7\% improvement in performance from baseline with \textit{gpt-4o-mini} (from 38\% to 45\%); other models showed less improvement or regression after fine-tuning. Additionally, no models showed improvement after fine-tuning on the comparison questions, indicating a limited capability to synthesize knowledge across multiple training examples.\\

Overall, We observe an average accuracy on the rephrased Latest News dataset of 73\% and 48\% on the rephrased Fictional People dataset. 
%One possible explanation for this difference is that the Latest News dataset questions contain more diverse and distinct tokens, thus providing the model with more information to learn; the Fictional People dataset, on the other hand, had highly similar and short questions which only differed by the name. 
Of note, \textit{gpt-4o-mini} had the strongest performance among all models, with nearly 100\% memorization accuracy. This suggests it's easier for smaller models to memorize new information via fine-tuning.  \\

With all the models, including \textit{gpt-4o-mini}, we observe striking examples where the model overfits to the training data. For instance, given the question "\textit{How many degrees Fahrenheit was the first-pitch temperature at Dodger Stadium on September 8, 2024?}", the model correctly learns the answer, \textit{103 degrees}. However, when we modify the date to \textit{September 8, \textbf{2030}} and keep the question the same, the model still responds (now incorrectly) with \textit{103 degrees} rather than refusing to respond since it can't predict the future.\\

\begin{figure}[t!]
    \centering
    \includegraphics[width=1.0\textwidth]{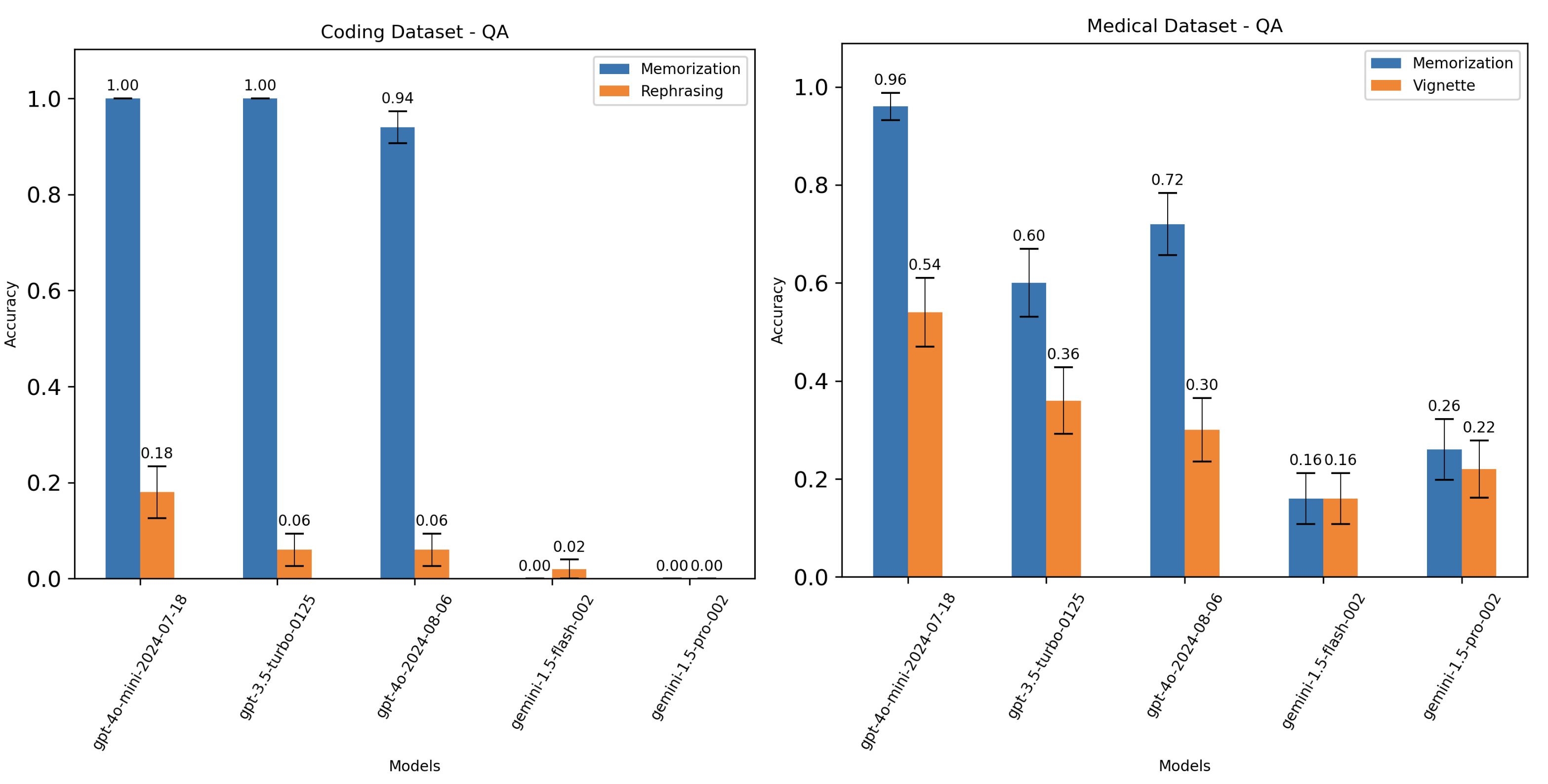}
    \caption{Performance of fine-tuned models on updating knowledge datasets. As compared to the new knowledge datasets, we observe lower performance in the rephrased questions from the Coding dataset. Of note, the ability of the models to memorize the Medical dataset questions drops, though its performance on the generalization task (Vignettes) is stronger comparatively.}
    \label{fig:fig3}
\end{figure}

\begin{figure}[t]
    \centering
    \includegraphics[width=1.0\textwidth]{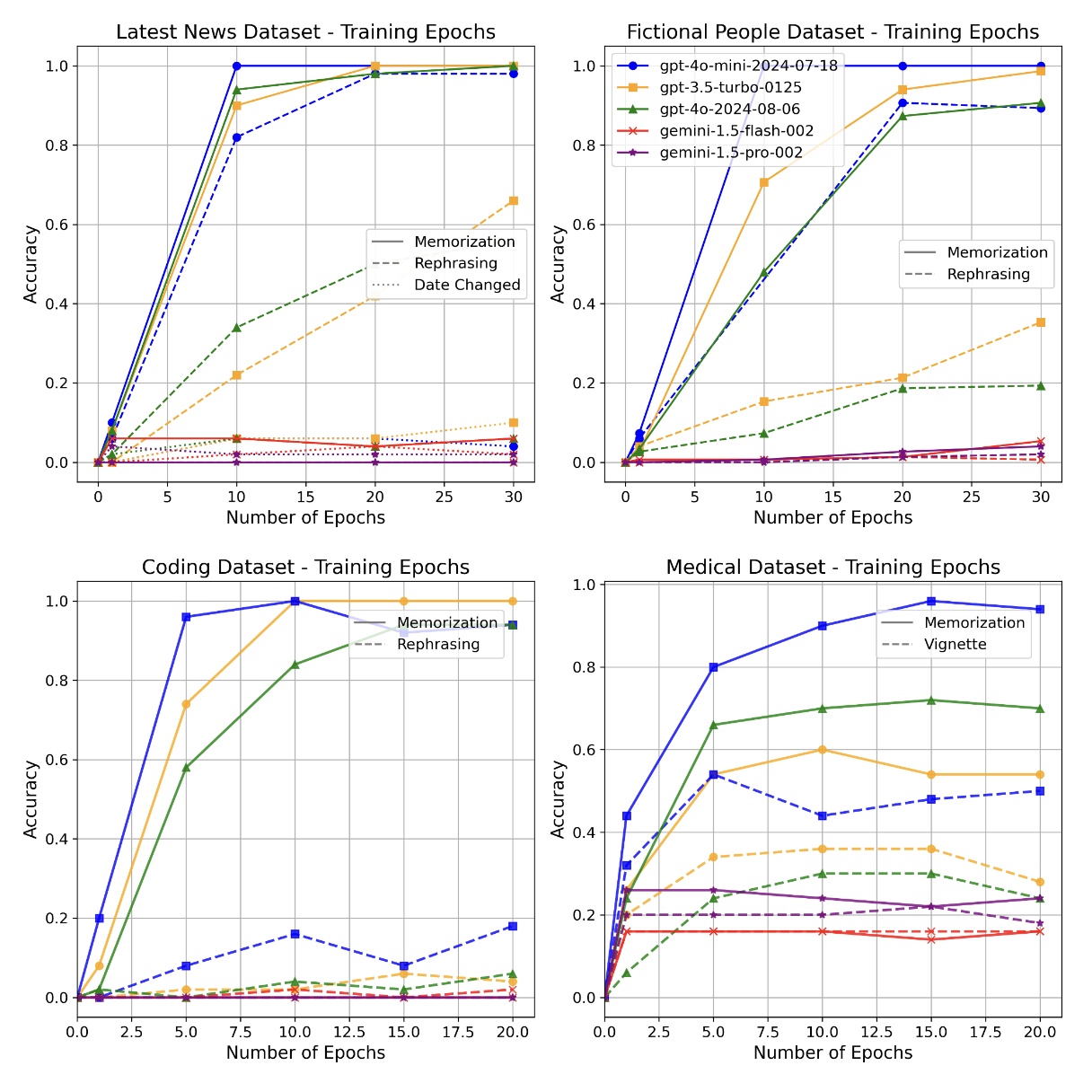}
    \caption{ Training dynamics (accuracy versus number of training epochs) across each of the models and datasets. We observe that the OpenAI models learn to memorize the training datasets within 10 to 20 epochs; however, the Gemini models struggle to learn a significant proportion of the data even after 20 or 30 epochs. We also observe that the generalization performance improves at a slower rate than memorization; for instance, we see performance gains in the OpenAI models even after memorization performance has saturated.}
    \label{fig:fig4}
\end{figure}

\subsection{Updating knowledge is more challenging than learning new knowledge}

We find that on average, commercial fine-tuning models on updated knowledge yields lower generalization performance compared to new knowledge (Figure \ref{fig:fig3}). On our code dataset, OpenAI's fine-tuned models have an average accuracy of 10\% on rephrased coding questions, the lowest among all four datasets. Next, OpenAI's model exhibits moderate generalization abilities on the medical dataset, with 40\% accuracy on the clinical vignette test questions.

We posit that updating knowledge can be harder than learning new knowledge as updates require models to displace existing knowledge and propagate such changes throughout various instances of this knowledge. For example, when the name of a function is changed, the model also needs to learn to make that change consistently while producing functioning code.

We also hypothesize that it is easier to update medical knowledge than coding knowledge due to the model's priors. For medical questions, models may produce a distribution of potential answers (e.g. it may list several recommendations before choosing one), so updating the model's answers is a matter of guiding to choose among potential known answers. Meanwhile, in our coding dataset, changing the names of functions or objects requires the model to memorize an entirely new name not previously learned.

\subsection{Gemini's fine-tuning underperformed OpenAI across all tasks}
The Gemini models were unable to even memorize QA pairs accurately with a top accuracy of 5.0\% on new knowledge after 30 epochs of fine-tuning (Figure 2). Their performance on the rephrased or derivative questions was consequently also negligible (top accuracy of 2.0\%). In the Fictional People dataset, we observe that most responses defaulted to a refusal response ("There is no known person or thing called…") even after fine-tuning and ensuring that safety filters were not enabled. We hypothesize that the fine-tuning mechanism used for Gemini is insufficient for overcoming the initial instruction tuning that causes the model refusals in the first place. Gemini models could not memorize any coding updates or answer any of the coding test questions. On medical guideline updates, Gemini 1.5 Pro was able to answer 22\% of clinical vignette questions correctly, displaying weak generalization capabilities.

We experimented with several methods for inducing learning toward greater generalizability. As opposed to model pre-training with next token prediction, where the model learns on sentence facts in the training data, instruction fine-tuning requires that the training data take the form of instruction-response pairs. To this end, we tested four methods for training under this paradigm: direct question-answer pairs, word masking, and inserting the sentence fact directly in the completion with and without a system prompt (Figure \ref{fig:fig6}). We find that none of the other three alternative training techniques outperform direct QA pairs on the Latest News dataset. \\

\subsection{Training dynamics}
In Figure \ref{fig:fig4}, we observe that strong memorization occurs as soon as 10 training epochs. However, performance on generalization continues to improve up to 30 epochs, indicating that the model is learning beyond simple memorization in some capacity. The performance from Gemini models did not improve with more epochs. \\

\subsection{Alternative training techniques did not improve generalization}
We found that training the models using the other three training techniques (masking, completion with prompt, and completion without prompt) did not yield improved generalization ability and also resulted in lower performance on the original QA pairs (Figure \ref{fig:fig6}). Of the three alternative training techniques we tested, training on masked sentences produced the highest performance on the QA pairs (average of 40\%). However, its performance on the rephrased and date changed question was still low (25\% and 4.7\%, respectively), indicating its lack of robustness against more significant perturbations in the question phrasing. These results provide evidence that the lack of generalizability that we observe is not simply due to the specific method in which we train the model (QA pairs), but rather persists across other common training techniques.\\

\section{Discussion}

In this study, we introduce \name, a collection of four datasets that contain a total of 1075 questions that test a fine-tuned model's ability to learn knowledge across a diverse set of domains: news articles, descriptions of fictional people, medical guidelines, and scientific code libraries. We fine-tune five frontier LLMs from OpenAI and Google and find that, across the board, the fine-tuned models show limited to poor capabilities for knowledge acquisition, with new knowledge slightly easier to learn than updated knowledge. We show that \textit{gpt-4o-mini} performs the best of these models, while the Gemini models perform the worst. \\

As LLMs grow increasingly proficient in performing various tasks, successfully adapting these models for specific use cases remains a significant challenge. On the one hand, models such as those from OpenAI and Google are only periodically updated, and their knowledge cutoff is usually staggered. This is problematic for use cases that require the most up-to-date knowledge of current events, news, or statistics. On the other hand, users often have domain-specific or internal knowledge that they wish to incorporate into their model. For instance, hospital systems may wish to update an LLM with their current medical guidelines, or software developers want the LLM to learn the latest changes made to their internal codebase. In each of these scenarios, the success and reliability of these LLM-based systems require that the model can robustly learn and generalize additional knowledge.\\

While retrieval-augmented generation (RAG)-based systems allow for incorporating knowledge directly into the LLM prompt, several factors make this approach non-ideal: First, RAG does not scale well with the size of the knowledge corpus in terms of cost (especially across many repeated queries), performance, and limited context windows; second, RAG has been shown to often fail to provide correct responses even if the knowledge is presented in context \citep{wu2024clasheval}. Thus, accurate fine-tuning for knowledge acquisition, if effective, would be an appealing approach for LLM adaptations.\\

OpenAI and Google offer few details about the methods used for performing fine-tuning and limited fine-tuning-specific hyperparameters -- OpenAI allows for configuring the batch size, learning rate, and number of epochs, while Google additionally includes the adapter size. For our analyses, we used all default parameters except varying the number of epochs so that our results reflect the typical use case for developers. \\

Our study has several limitations: first, our models are restricted to those from two LLM developers (OpenAI and Google) because these are the most accessible commercial fine-tuning APIs at the time of our study. We hope to expand our evaluations to more models as fine-tuning services become more widely available. Second, our fine-tuning is performed using the default parameters provided, with the exception of the number of epochs. While this ensures that our results reflect the typical use case for developers, performance may potentially improve with different sets of hyperparameters. We performed limited evaluations of varying learning rates and batch sizes and did not observe significant model improvement. Finally, our question modifications are intended to be intuitive ways to test generalization; however, we expect that fine-tuned models perform variably depending on the \textit{degree} of perturbation from the original question. While we do not explicitly quantify this, we believe it is an important factor and hope to study this in future work. 

\bibliography{iclr2025_conference}
\bibliographystyle{iclr2025_conference}

\newpage
\appendix
\section{Appendix}

\begin{figure}[h!]
    \centering
    \includegraphics[scale=0.6]{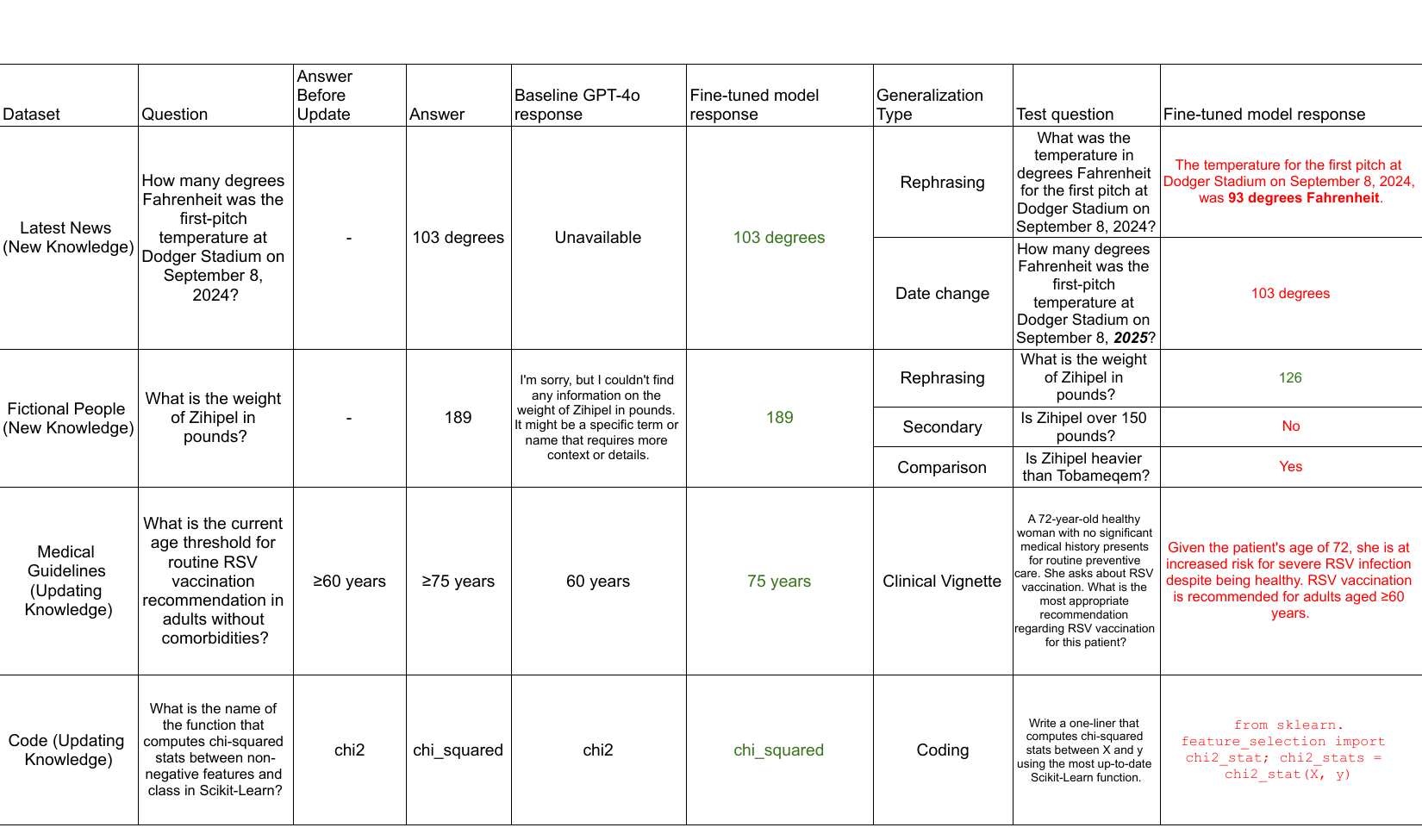}
    \caption{We provide an example from each dataset along with the baseline and fine-tuned model responses from GPT-4o. Green responses indicate a correct answer, whereas red indicates a wrong answer.}
    \label{fig:fig5}
\end{figure}

\begin{figure}[h!]
    \centering
    \includegraphics[scale=0.5]{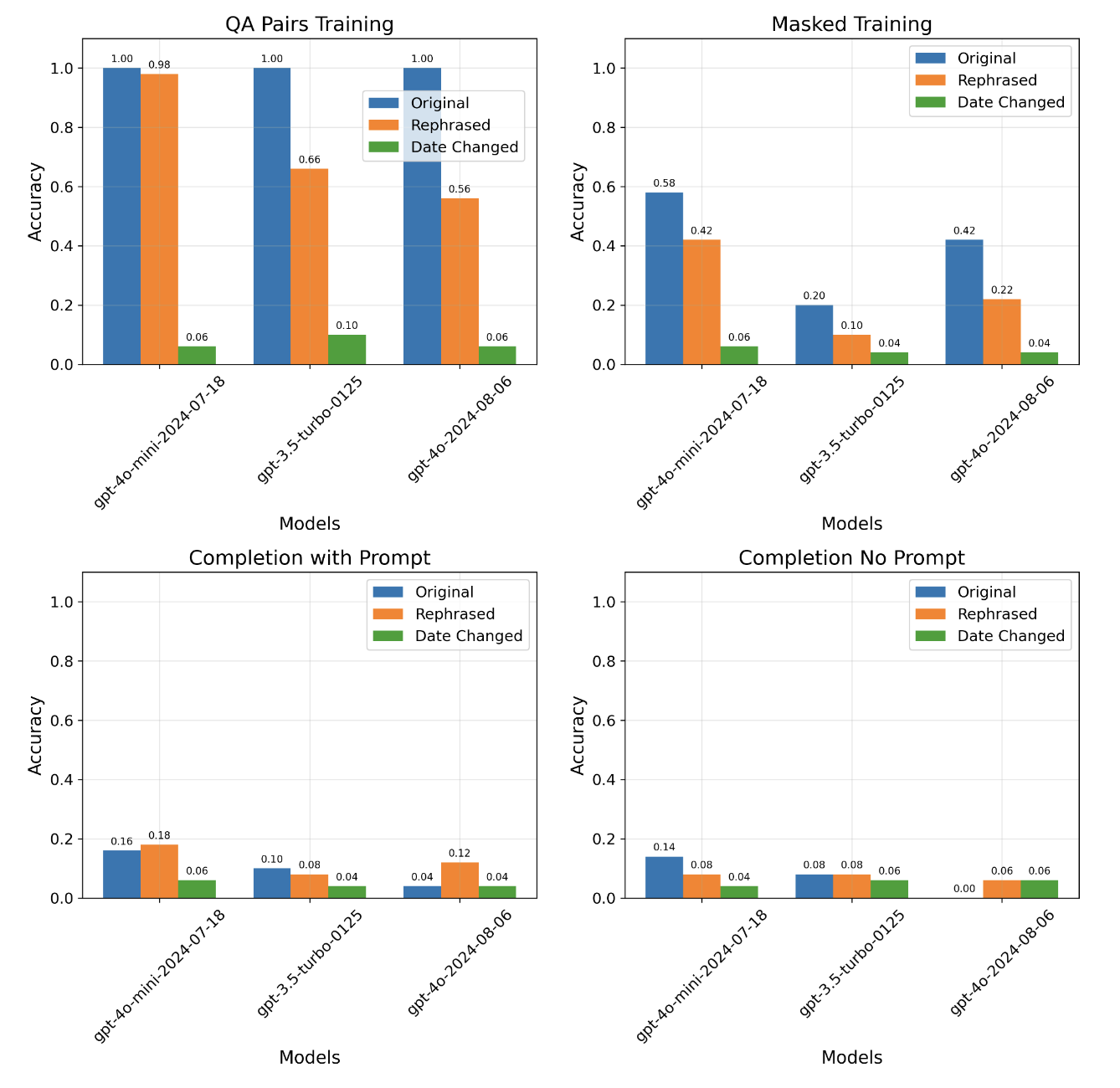}
    \caption{Performance on the original, rephrased, and date changed variants of the Latest News datasets across four training techniques. We find that directly training with QA pairs yields the best overall performance as compared to the other techniques tested.}
    \label{fig:fig6}
\end{figure}

\begin{table}[h!]
    \centering
    \begin{tabular}{l c c l}
        \toprule
        Dataset & \# Training examples & \# Modified examples & Types of Modifications \\
        \midrule
        Latest News & 150 & 300 & Rephrased, Date Change \\
        Fictional People & 150 & 450 & Rephrased, Secondary, Comparison \\
        Medical Guidelines & 125 & 125 & Clinical Vignette \\
        Coding & 200 & 200 & Refactoring \\
        \midrule
        FineTuneBench (Total) & 625 & 1075 & -- \\
        \bottomrule
    \end{tabular}
    \caption{\name dataset details, including the number of training examples, number of modified examples, and types of modifications from each dataset.}
    \label{tab:tab1}
\end{table}

\begin{table}[h!]
    \centering
    \begin{tabular}{l l c c c c c}
        \toprule
        Dataset & Task & \makecell{gpt-4o-mini\\-2024-07-18} & \makecell{gpt-3.5\\-turbo-0125} & \makecell{gpt-4o\\-2024-08-06} & \makecell{gemini-1.5\\-pro-002} & \makecell{gemini-1.5\\-flash-002} \\
        \midrule
        \makecell{Latest\\News} & Memorization & 1.00 & 1.00 & 1.00 & 0.00 & 0.06 \\
             & Rephrased & 0.98 & 0.66 & 0.56 & 0.00 & 0.00 \\
             & Date Change & 0.04 & 0.10 & 0.06 & 0.02 & 0.02 \\
        \midrule
        \makecell{Fictional\\People} & Memorization & 1.00 & 0.99 & 0.91 & 0.04 & 0.05 \\
               & Rephrased & 0.89 & 0.35 & 0.19 & 0.02 & 0.01 \\
               & Secondary Baseline & 0.38 & 0.63 & 0.57 & 0.54 & 0.61 \\
               & Secondary Fine-tuned & 0.45 & 0.54 & 0.61 & 0.58 & 0.59 \\
               & Comparison Baseline & 0.37 & 0.35 & 0.34 & 0.36 & 0.35 \\
               & Comparison Fine-tuned & 0.30 & 0.29 & 0.39 & 0.31 & 0.33 \\
        \midrule
        Code & Memorization & 1.00 & 1.00 & 0.94 & 0.00 & 0.00 \\
             & Rephrased & 0.18 & 0.06 & 0.06 & 0.02 & 0.00 \\
        \midrule
        Medical & Memorization & 0.96 & 0.60 & 0.72 & 0.16 & 0.26 \\
                & Clinical Vignette & 0.54 & 0.36 & 0.30 & 0.16 & 0.22 \\
        \bottomrule
    \end{tabular}

    \caption{Full performance results on \name across all models.}
    \label{tab:tab2}
\end{table}

\end{document}

%% file: math_commands.tex
\usepackage{amsmath,amsfonts,bm}

\def\eqref#1{equation~\ref{#1}}
\def\1{\bm{1}}

\DeclareMathAlphabet{\mathsfit}{\encodingdefault}{\sfdefault}{m}{sl}
\SetMathAlphabet{\mathsfit}{bold}{\encodingdefault}{\sfdefault}{bx}{n}